\documentclass{llncs}
\usepackage{graphicx}
\usepackage{xcolor}
\usepackage{url}
\makeatletter
\g@addto@macro{\UrlBreaks}{\UrlOrds}
\makeatother

\begin{document}

\title{An Ethical Black Box for Social Robots: a draft Open Standard}
\titlerunning{EBB Open Standard}  
%
\author{Alan F.T. Winfield\inst{1}\thanks{alan.winfield@brl.ac.uk} \and Anouk van Maris\inst{1} \and Pericle Salvini\inst{2} \and Marina Jirotka\inst{2}}
\institute{Bristol Robotics Lab, University of the West of England, Bristol
\and {Department of Computer Science, University of Oxford}}
%
%
\tocauthor{Alan Winfield}

\maketitle              

\begin{abstract}
This paper introduces a draft open standard for the robot equivalent of an aircraft flight data recorder, which we call an ethical black box. This is a device, or software module, capable of securely recording operational data (sensor, actuator and control decisions) for a social robot, in order to support the investigation of accidents or near-miss incidents. The open standard, presented as an annex to this paper, is offered as a first draft for discussion within the robot ethics community. Our intention is to publish further drafts following feedback, in the hope that the standard will become a useful reference for social robot designers, operators and robot accident/incident investigators.
\end{abstract}
\section{Introduction}
In \cite{winf2017} we argued the case that robots and autonomous systems should be equipped with the equivalent of an aircraft Flight Data Recorder to continuously record sensor and relevant internal status data. We call this an ethical black box (EBB). We argued that an ethical black box will play a key role in the processes of discovering why and how a robot caused an accident, and thus an essential part of establishing  accountability and responsibility.

We propose that the EBB needs a standard specification. A standard specification has several benefits. First, a standard approach to EBB implementation in social robots will greatly benefit accident and incident (near miss) investigations \cite{winf2020b}. Second, an EBB will provide social robot designers and operators with data on robot use that can support both debugging and functional improvements to the robot. Third, an EBB can be used to support robot `explainability' functions to allow, for instance, the robot to answer `Why did you just do that?' questions from its user. And fourth, a standard allows EBB implementations to be readily shared and adapted for different robots and, we hope, encourage manufacturers to develop and market general purpose robot EBBs. 

This paper is structured as follows. Section 2 provides a brief recap of the history of data loggers, and associated standards, in aviation, critical infrastructure and road vehicles. In section 3 we give a brief high-level description of the EBB, and in section 4 we argue the case for a standardised EBB, and a draft open standard as a starting point. In section 5 we conclude the paper by outlining the draft open standard in Annex A.

\section{A brief introduction to Data Loggers}

The term `black box' was first used informally in the late 1940s for navigational instruments, within the Royal Air Force. The term was then extended to cover any kind of apparatus within a sealed container. From the mid 1960s the colloquial use of the `black box' has narrowed to refer to the Flight Data Recorder (FDR), now fitted as standard in aircraft.

Black box – or flight data recorders – were introduced in 1958, for larger aircraft, and since then have vastly expanded in scope in what flight data they record. Initially FDRs included time navigation data about the position of surfaces and the pilots' movement of controls; latterly sensor data on the internal and external environment as well as the functioning of components and systems are also recorded, alongside autopilot settings such as selected headings, speeds, altitudes and so on \cite{Gross2006}. FDRs on modern aircraft record more than 1000 parameters \cite{NTSB2002}. 

The transfer of the black box concept into settings other than aviation is not new. Data loggers for critical infrastructure such as Supervisory Command and Data Acquisition (SCADA) systems are also standard practice \cite{Morr2010,Oh2018}. The largest deployment of black box technology outside aviation is within the automobile and road haulage industries for data logging \cite{Thom2008,Worr2016}. Data loggers for vehicles are generally known as Event Data Recorders (EDRs). Standards for EDRs include IEEE 1616 `Standard for Motor Vehicle Event Data Recorder (MVEDR)' first published in 2004 and revised in 2021 \cite{IEEE1616}.

\section{The Ethical Black Box}

\begin{figure}[ht!]
	\begin{center}
		\includegraphics[width=11cm]{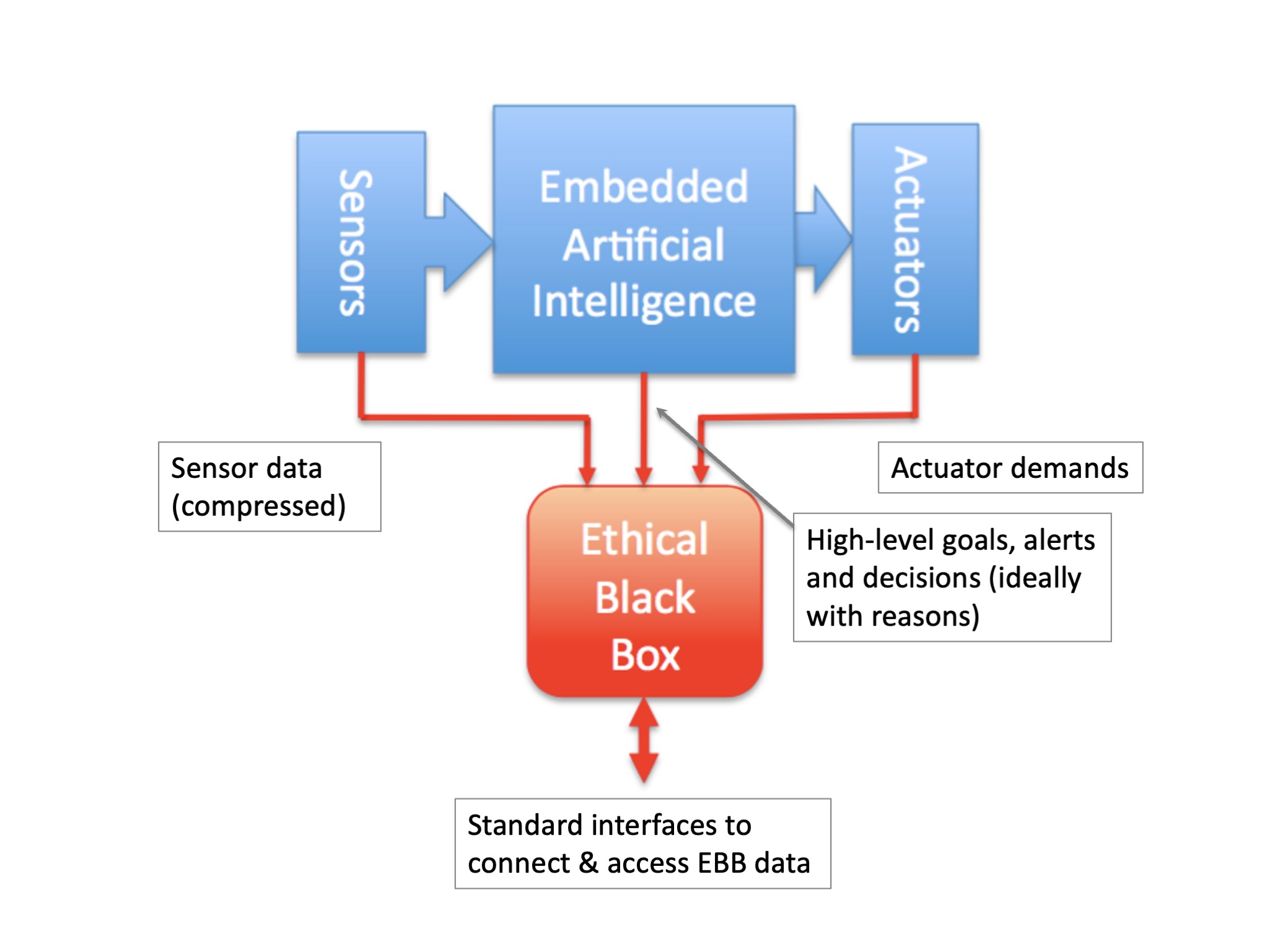}
        
 		\textbf{\refstepcounter{figure}\label{fig:02} Figure \arabic{figure}. }{Robot sub-systems with an Ethical Black Box and key dataflows.}
    \end{center}
\end{figure}

All robots collect sense data, and -- on the basis of that sense data and some internal decision making process (AI) -- send commands to actuators. This is of course a simplification of what in practice will be a complex set of connected systems and processes but, at an abstract level, all intelligent robots will have the three major subsystems shown in blue, in Fig. \ref{fig:02}. A social robot is no different, except that it is designed to interact directly with humans. 

We define the ethical black box (EBB) as a system for securely recording date- and time-stamped operational data from a social robot. We use the term \textit{ethical black box} to emphasise our view that to deploy social robots without an EBB would be irresponsible.

The Ethical Black Box (EBB) and its data flows, shown in red in Fig. \ref{fig:02} will need to collect and store data from all three robot subsystems: sensor data, actuator demands and actual positions, and robot decisions --- ideally with the reasons for those decisions. All of these data will need to be date and time stamped. An important property of the EBB is that the data flows from the robot to the EBB are strictly one-way. It is important that the EBB is a passive sub-system, accepting data from the robot controller, and not affecting the robot's operation.

\section{Why do we need a standardised EBB?}

We contend that social robots should not only be fitted with an EBB, but that EBB should follow a standard specification. There are several benefits from a standard EBB:

\begin{enumerate}
\item A standard approach to EBB implementation in social robots will greatly benefit accident and incident (near-miss) investigations. In \cite{winf2020b} we argue that social robots bring greater risks than industrial robots, and hence the likelihood of harms --- including psychological, societal or environmental harms --- is greater. Without the data provided by the EBB the investigation of accidents or near-miss incidents in order to discover what happened, why it happened and how to prevent it happening again is difficult, if not impossible.
\item An EBB will provide social robot designers and operators with data on robot use that can support both debugging and functional improvements to the robot. Thus it becomes a powerful diagnostic tool during research and development.
\item An EBB can be used to support robot explainability functions to allow, for instance, the robot to answer ``Why did you just do that?” questions from its user \cite{Koem2020}.\footnote{Noting that this would require the robot's control system to access it's own EBB - and thus violate the principle of the EBB's passivity.} And
\item a standard allows EBB implementations to be readily shared and adapted for different robots and, we hope, both encourage manufacturers to develop and market general purpose robot EBBs, and regulators to require EBBs.
\end{enumerate}

Bruce Perens, creator of The Open Source Definition, outlines a number of criteria an open standard must satisfy, including:

\begin{enumerate}
\item ``Availability: Open standards are available for all to read and implement.
\item Maximize End-User Choice: Open Standards create a fair, competitive market for implementations of the standard. 
\item No Royalty: Open standards are free for all to implement, with no royalty or fee. 
\item No Discrimination: Open standards and the organizations that administer them do not favor one implementor over another for any reason other than the technical standards compliance of a vendor's implementation.
\item Extension or Subset: Implementations of open standards may be extended, or offered in subset form.''\footnote{https://opensource.com/resources/what-are-open-standards}
\end{enumerate}

\section{The Draft Open Standard}

In Annex A we set out the first draft of an Open Standard for an EBB for social robots. Following the model of the Internet open standards this draft is a Request for Comments (RFC). Subsequent drafts will incorporate feedback. The standard is written following BS 0:2021 \textit{A Standard for Standards} \cite{BS0}.

Annex A sets out the requirements for an EBB Software Module that could be integrated within a robot's controller, or as a stand alone software process connected, via a network connection, to the robot.

Annex A details the normative requirements for the data structures and formatting. These data structures fall into three categories: Meta Data, which stores information about the robot that the EBB is connected to, Data Data, which stores information on the number of records and dates and times of the oldest and most recent records in the EBB, and Robot Data, which stores operational data on the robot. The final category Robot Data will comprise most of the data stored in the EBB. All three structures are date- and time-stamped and have a common format. Annex A also provides examples of records for each of these three data structures together with an example of a complete set of EBB records, and how often particular robot data records might be written.

The overall aim of the standard is to provide a technical specification for an EBB for Social Robots. This specification, alongside an open-source library of model implementations, will provide a resource to enable developers to build an EBB into their robots.

\section*{Acknowledgments}
This work has been conducted within project RoboTIPS: Developing Responsible Robots for the Digital Economy supported by EPSRC grant ref EP/S005099/1.

\bibliographystyle{plain}
\bibliography{EBBpaper}

\section*{Annex A: Draft Standard for an Ethical Black Box Software Module for Social Robots}

\section*{Request For Comments, Draft 0.1}

\subsection*{A.1 Scope}

This draft open standard sets out normative technical requirements for a data logger for Social Robots, which we call an Ethical Black Box (EBB). This draft is a Request For Comments (RFC), inviting feedback and suggestions for improvements. Further drafts will incorporate revisions in response to this feedback.

The aim of the standard is to provide researchers, developers and operators with a technical specification for an EBB for Social Robots. This specification, alongside an open source library of model implementations, will provide a resource to enable developers to build an EBB into their robots.

The standard sets out the specification for a software module for an EBB, which may be implemented either as a software module alongside the robot's control system, or as a stand alone software process within a system connected (i.e. via a network connection) to the robot. This module may also be implemented within a hardware EBB physically connected to the robot, although the specification of the hardware is outside the scope of this draft.

\subsection*{A.2 Normative References}

\begin{itemize}
\item BS 0:2021, \textit{A standard for standards — Principles of standardization}
\item BS ISO 8373:2021, \textit{Robotics — Vocabulary}
\end{itemize}

\subsection*{A.3 Terms and Definitions}

For the purposes of this document, the following terms and definitions apply. 

\textbf{Ethical Black Box (EBB)}: A system for securely recording date- and time-stamped operational data from a social robot.

\textbf{Social Robot}: An intelligent service robot  designed to interact with humans. For definitions of intelligent service robot refer to BS ISO 8373:2021. 

\subsection*{A.4 EBB Normative Requirements for Data}

This section sets out requirements for the data that should be captured by the EBB. 

\paragraph{\textbf{EBB data organisation}}

The data stored in the EBB shall comprise three types of record: meta data, data data and robot data. The EBB shall contain one meta data (MD) record only, one data data (DD) record only, and a number of robot data (RD) records. The maximum number of RD records shall be fixed for each EBB according to the limits of its storage capacity. The general organisation of EBB data is shown in Table \ref{table1}.

\begin{table}[h]
\centering
    \begin{tabular}{|l|} 
    \hline
    EBB records \\
    \hline\hline
    Meta data (MD) record \\ 
    \hline
    Data data (DD) record \\
    \hline
    Robot data (RD) record 1 \\ 
    \hline
    Robot data (RD) record 2 \\ 
    \hline
    ... \\ 
    \hline
    Robot data (RD) record $n$ \\ 
    \hline
    \end{tabular}
    \caption{General organisation of EBB data}
	\label{table1}
\end{table}

The EBB RD logs shall be written in order starting from RD record 1, and proceeding to RD record $n$. After record $n$ has been written the EBB shall write the next RD record overwriting RD record 1, and write subsequent RDs in record 2, and so on. In this way the EBB always stores the most recent set of RD records, up to its maximum capacity of $n$ records.

Meta Data, Data data and Robot Data records have the same overall structure, as shown in \ref{table2}.

\begin{table}[h]
\centering
    \begin{tabular}{|l|c|c|} 
    \hline
    Record & length & format \\
    \hline\hline
    Record type `MD', `DD' or `RD' & 2 chars & ASCII text \\ 
    \hline
    Number of fields and chars in record & 12 chars & ASCII 000:00000000 \\ 
    \hline
    Date record written & 10 chars & ASCII yyyy:mm:dd \\
    \hline
    Time record written & 12 chars & ASCII hh:mm:ss:ms \\
    \hline
    Data record 1 & variable & see below \\ 
    \hline
    ... & &\\
    \hline 
    Data record $m$ & variable & see below \\ 
    \hline
    Checksum & 4 chars & ASCII \\
    \hline
    \end{tabular}
    \caption{Common structure of Meta Data, Data Data and Robot Data Records}
	\label{table2}
\end{table}

Each EBB record consists of a 2 character record labels `MD', `DD' or `RD' followed by a variable number of fields.

Each field in a record consists of a 4 character label, followed by data elements defined according to the label. EBB fields are defined below for each of the 3 types of record. Note that several fields are common to all 3 record types.

\paragraph{\textbf{The EBB Meta Data Record}}

The Meta Data Record shall store information about the robot that the EBB is fitted to (name, version or model no, and serial no), the robot's developer/manufacturer and operator, the contact details of the person responsible for the robot, and information on the EBB itself. 

Table \ref{table3} defines each field in the MD record, and includes both required and optional fields. Here a string is defined as a variable length ASCII sequence terminated by the null character ASCII \textbackslash0.

\begin{table}[h]
\centering
    \begin{tabular}{|c|l|c|c|} 
    \hline
    label & data & length & requirement \\
    \hline\hline
    recS & record size, field and chars, including recS field & 12 chars & required \\
    \hline
    ebbD & EBB date record written & 10 chars & required \\
    \hline
    ebbT & EBB time record written & 12 chars & required \\
    \hline
    botN & robot name & string & required \\
    \hline
    botV & robot version no & string & optional \\
    \hline
    botS & robot serial no & string & optional \\
    \hline
    botM & robot manufacturer & string & required \\
    \hline
    opeR & robot operator & string & optional \\
    \hline
    resP & name and contact details of responsible person & string & required \\
    \hline
    ebbN & EBB name and version no & string & required \\
    \hline
    chkS & checksum for complete record & 8 hexadecimal chars & required \\
    \hline
    \end{tabular}
    \caption{Meta Data Fields}
	\label{table3}
\end{table}

Notes on the MD fields:
\begin{enumerate}
\item The record size \emph{recS}, has a 12 char data element formatted as 3 numeric chars and 8 numeric chars separated by a colon, i.e. 000:00000000. It follows that the maximum permissible number of fields in a record is 999, and the maximum number of characters is 99,999,999. Note also that the size and character count must include the recS field.
\item The EBB date field \emph{ebbD} is 12 ASCII characters with colon separated year, month, and day, i.e. yyyy:mm:dd.
\item The EBB time field \emph{ebbT} is 10 ASCII characters with colon separated hour, minute, second and millisecond, i.e. hh:mm:ss:ms.
\item The record checksum \emph{chkS} shall be computed using a 64-bit non-cryptographic hash function, to be determined. 
\end{enumerate}

An example of a complete Meta Data Record is shown in \ref{table4}.

\begin{table}[h]
\centering
    \begin{tabular}{|l|l|} 
    \hline
    field & comment \\
    \hline\hline
    MD & record label\\
    \hline
    recS 010:00000000 & number of fields:chars in record \\
    \hline
    ebbD 2022:04:20 & date 20 April 2022 \\
    \hline
    ebbT 16:40:20:000 & time 16:40 and 20.000 seconds \\
    \hline
    botN NAO\textbackslash0 & NAO robot \\
    \hline
    botV 4\textbackslash0 & v4 \\
    \hline
    botM Aldebaran\textbackslash0 & Manufacturer \\
    \hline
    opeR Bristol Robotics Lab\textbackslash0 & Operator \\
    \hline
    resP A Winfield +44 117 328 6913\textbackslash0 & person responsible \\
    \hline
    ebbN PyEBB v1.2\textbackslash0 & this EBB \\
    \hline
    chkS AF5679FC & checksum for this record \\
    \hline
    \end{tabular}
    \caption{An example Meta Data Record}
	\label{table4}
\end{table}

\paragraph{\textbf{The EBB Data Data Record}}

The Data Data Record shall store information about the robot data records stored in the EBB.

Table \ref{table3} defines each field in the DD record, and includes both required and optional fields.

\begin{table}[h]
\centering
    \begin{tabular}{|c|l|c|c|} 
    \hline
    label & data & length & requirement \\
    \hline\hline
    recS & record size, field and chars & 12 chars & required \\
    \hline
    ebbN & total number of EBB Data Records stored in EBB & 10 chars & required \\
    \hline
    ebbX & index to the start of next writable RD record & 16 chars & required \\
    \hline
    ebD1 & date of oldest RD record written & 10 chars & required \\
    \hline
    ebT1 & time of oldest RD record written & 12 chars & required \\
    \hline
    ebDM & date of most recent RD record written & 10 chars & required \\
    \hline
    ebTM & time of most recent RD record written & 12 chars & required \\
    \hline
    sysX & manufacturer definable field & variable & optional\\
    \hline
    chkS & Checksum for complete record & 8 hexadecimal chars & required \\
    \hline
    \end{tabular}
    \caption{Data Data Fields}
	\label{table5}
\end{table}

Notes on the DD fields:
\begin{enumerate}
\item For notes on \emph{recS} \emph{ebD1}, \emph{ebT1}, \emph{ebDM}, \emph{ebTM} and \emph{chkS} see notes on Table \ref{table3} above.
\item Field \emph{ebbX} is an offset, in number of bytes, from the start of the EBB storage media to the next writable position for an RD record. Note that this will need to be reset back to RD 1 once the storage media is full.
\item Field \emph{sysX} `system exclusive' is  manufacturer/operator definable. The data is formatted as 2 chars and a string separated by a colon, i.e. 00:string. The 2 characters are to allow the manufacturer to define up to 99 sysX fields, and the string allows for variable length data.
\item Given that all fields in the DD record are required and have a fixed length the \emph{recS} field will have the default value 010:00000130, as shown in the example DD record in Table \ref{table6}. Only if the record includes sysX field(s) will chkS have a different value.
\end{enumerate}

An example of a complete Data Data Record is shown in \ref{table6}.

\begin{table}[h]
\centering
    \begin{tabular}{|l|l|} 
    \hline
    field & comment \\
    \hline\hline
    DD & record label\\
    \hline
    recS 010:000000130 & number of fields:chars in record \\
    \hline 
    ebbD 2022:04:20 & date 20 April 2022 \\
    \hline
    ebbT 16:40:20:000 & time 16:40 and 20.000 seconds \\
    \hline
    ebbN 0000000400 & 400 records in EBB \\
    \hline
    ebbX 00000000001545060 & offset to next RD position in storage media \\
    \hline
    ebD1 2022:03:01 & first RD date 1 March 2022 \\
    \hline
    ebT1 08:00:30:000 & first RD time 08:00 and 30.000 seconds \\
    \hline
    ebDM 2022:05:01 & Most recent RD date 1 May 2022 \\
    \hline
    ebTM 18:59:30:100 & Most recent RD time 18:59 and 30.100 seconds \\
    \hline
    chkS FF5678AC & checksum for this record \\
    \hline
    \end{tabular}
    \caption{An example Data Data Record}
	\label{table6}
\end{table}

\paragraph{\textbf{EBB Robot Data Records}}

The Robot Data Records store operational data from the robot

Table \ref{table7} defines each field in the RD record, and includes both required and optional fields.

\begin{table}[h]
\centering
    \begin{tabular}{|c|l|c|c|} 
    \hline
    label & data & length & requirement \\
    \hline\hline
    botT & robot time & 10 chars & required \\
    \hline
    actD & actuator no and demand value & 12 chars 000:±0000.00 & optional \\
    \hline
    actV & actuator no and actual value & 12 chars 000:±0000.00 & optional \\
    \hline
    batL & battery level & 3 chars & optional \\
    \hline
    tchS & touch sensor no and value & 6 chars 00:000 & optional \\
    \hline
    irSe & infra red sensor no and value & 6 chars 00:000 & optional \\
    \hline
    lfSe & line following sensor no and value & 6 chars 00:000 & optional \\
    \hline
    gyrV & gyro no and value & 20 chars 00:±0000:±0000:±0000 & optional\\
    \hline
    accV & accelerometer no and value & 20 chars 00:±0000:±0000:±0000 & optional\\
    \hline
    tmpV & temperature sensor no and value & 8 chars 00:±0000 & optional \\
    \hline
    micI & microphone no and input & variable, 2 chars:8 chars:wav hex & optional \\
    \hline
    camF & camera no and frame grab & variable, 2 chars:8 chars:jpg hex & optional \\ 
    \hline
    txtC & text input command & variable, string & optional \\
    \hline
    txtR & text reply & variable, string & optional \\
    \hline
    decC & robot decision code and reason & variable, 4 chars 0000:string & optional \\
    \hline
    wifi & WiFi status and signal strength &
    4 chars 0:00 & optional \\
    \hline
    sysX & manufacturer definable field & variable, 2 chars 00:string & optional\\
    \hline
    chkS & checksum for complete record & 8 hexadecimal chars & required \\
    \hline
    \end{tabular}
    \caption{Robot Data Fields}
	\label{table7}
\end{table}

Notes on the RD fields:
\begin{enumerate}
\item For notes on \emph{recS}, and \emph{chkS} see notes on Table \ref{table3} above. For notes on \emph{sysX} see notes on Table \ref{table5}.
\item Field \emph{botT} is needed in case the robot's clock shows a different time to the EBB's clock. The format of \emph{botT} is the same as \emph{ebbT}: 10 ASCII chars, for hours, minutes, seconds and milliseconds 00:00:00:000.
\item Fields \emph{actD} and \emph{actV} each contain values for the actuator number and the actuator demand, or actual positions respectively. The data is formatted 000:±0000.00 thus allowing for a maximum of 999 actuators, and positive or negative values of up to ±9999.99.
\item Fields \emph{tchS, irSe, lfSe} each contain values for sensor number and the sensor value. The data is formatted 00:000 thus allowing for up to 99 sensors, and sensor values between 0...999.
\item Field \emph{micI} allows for storage of a wav audio clip captured by the robot's microphone(s). The data has three colon separated elements, 2 chars for the microphone number, 8 chars for the length of the wav clip, and the hexadecimal representation of the wav binary. This allows for up to 99 microphones and wav clips of up to 99,999,999 bytes. 
\item Field \emph{camF} allows for storage of a jpg still frame grabbed by the robot's camera(s). The data has three colon separated elements, 2 chars for the camera number, 8 chars for the length of the jpg clip, and the hexadecimal representation of the jpg binary. This allows for up to 99 cameras and jpg clips of up to 99,999,999 bytes.
\item Fields \emph{txtC} and \emph{txtR} allow for storage of text input commands to the robot, and text responses from the robot, respectively. The input command might be typed by the robot's user, or spoken and then converted from speech to text by the robot's speech recognition system. The output text might be displayed visually or spoken by the robot's speech synthesis system. The data elements for both \emph{txtC} and \emph{txtR} are null terminated strings.
\item Field \emph{decC} allows for storage of the robot's internal decision of its next action (i.e. turn left, turn right, stop, speak an alert, etc), together (optionally) with a reason for that decision. The data is formatting as a 4 char decision code, followed by a string. This allows for up to 9999 decisions to be logged. The determination of which numeric value to use for each robot decision is outside the scope of this standard, and left to the robot's manufacturer or operator. If there is no reason the string shall be stored as an empty null string terminated by ASCII \textbackslash0.
\item Field \emph{wifi} allows for storage of the robot's WiFi connection status and signal strength. The colon separated data is formatted as 1 character connection status, connected (1) or not connected (0), and 2 characters for signal strength, from 00..99.
\end{enumerate}

An example of a complete Robot Data Record is shown in Table \ref{table6}, for a simple differential drive wheeled robot, with 8 IR sensors.

\begin{table}[h]
\centering
    \begin{tabular}{|l|l|} 
    \hline
    field & comment \\
    \hline\hline
    RD & record label\\
    \hline
    recS 017:00000nnn & number of fields:chars in record \\
    \hline
    ebbD 2022:04:20 & RD date \\
    \hline
    ebbT 16:40:20:000 & Rd time \\
    \hline
    batL 255 & battery level \\
    \hline
    actV 001:-175.54 & left wheel angle \\
    actV 002:102.09 & right wheel angle \\
    \hline
    irSe 001:0.05 & IR sensor 1 \\
    irSe 002:0.05 & IR sensor 2 \\
    irSe 003:0.05 & IR sensor 3 \\
    irSe 004:0.05 & IR sensor 4 \\
    irSe 005:0.05 & IR sensor 5 \\
    irSe 006:0.23 & IR sensor 6 \\
    irSe 007:0.15 & IR sensor 7 \\
    irSe 008:0.05 & IR sensor 8 \\
    \hline
    decC 0020:obstacle detected\textbackslash0 & turning left to avoid obstacle on right \\
    \hline
    wifi 1:255 & WiFi connected, good signal \\
    \hline
    chkS CFA3569A & checksum for this record \\
    \hline
    \end{tabular}
    \caption{An example Robot Data Record}
	\label{table8}
\end{table}

\subsubsection*{A.4.2 EBB Timing}

Not all EBB records need to be written with the same frequency. In general those RD records which capture actuator movements and short range sensor inputs will need to be written with the highest frequency; a default setting for these might be once every 2 seconds. RD records that capture camera images might be written with a lower frequency, especially if the robot moves slowly, say once every 10 seconds, or less. A third group of RD records are those that capture aperiodic events, such as a user's commands and the robot's response (if there is one). 

Table \ref{table9} illustrates the EBB timing, with three kinds of event logged: high frequency motor and sensor values captured once every 2 seconds (in RD 1, 2, 4, 6 etc), lower frequency camera frame grabs once every 10 seconds (in RD 3, RD 10 and RD 17), and the sporadic events of user commands to override the robot's autonomous operation, in RD 5 and RD 13. For clarity only the \emph{ebbT} time fields are shown in full. Abbreviated \emph{camF} and \emph{txtC} fields are also shown in RD 3, 5, 10, 13 and 17.

\begin{table}[h]
\centering
    \begin{tabular}{|l|l|} 
    \hline
    label & fields \\
    \hline\hline
    MD & recS... ebbD... ebbT 08:40:20:000 BotN ePuck BotM... Resp... chkS...\\ 
    \hline
    DD & recS... ebbD... ebbT 08:40:20:000 ebbD1... ebbT1... ebbN...  ebDM... ebbTM... chkS... \\
    \hline 
    RD 1 & recS... ebbD... ebbT 08:40:22:000 botT... actV... batL... irSE... chkS...\\ 
    \hline
    RD 2 & recS... ebbD... ebbT 08:40:24:000 botT... actV... batL... irSE... chkS...\\ 
    \hline  
    RD 3 & recS... ebbD... ebbT 08:40:25:000 botT... camF 01:00307200:... chkS...\\ 
    \hline
    RD 4 & recS... ebbD... ebbT 08:40:26:000 botT... actV... batL... irSE... chkS...\\ 
    \hline   
    RD 5 & recS... ebbD... ebbT 08:40:27:100 botT... txtC Halt\textbackslash0 chkS...\\ 
    \hline
    RD 6 & recS... ebbD... ebbT 08:40:28:000 botT... actV... batL... irSE... chkS...\\ 
    \hline 
    RD 7 & recS... ebbD... ebbT 08:40:30:000 botT... actV... batL... irSE... chkS...\\ 
    \hline
    RD 8 & recS... ebbD... ebbT 08:40:32:000 botT... actV... batL... irSE... chkS...\\ 
    \hline  
    RD 9 & recS... ebbD... ebbT 08:40:34:000 botT... actV... batL... irSE... chkS...\\ 
    \hline   
    RD 10 & recS... ebbD... ebbT 08:40:35:000 botT... camF 01:00307200:... chkS...\\ 
    \hline
    RD 11 & recS... ebbD... ebbT 08:40:36:000 botT... actV... batL... irSE... chkS...\\ 
    \hline
    RD 12 & recS... ebbD... ebbT 08:40:38:000 botT... actV... batL... irSE... chkS...\\ 
    \hline
    RD 13 & recS... ebbD... ebbT 08:40:27:100 botT... txtC Run\textbackslash0 chkS...\\ 
    \hline
    RD 14 & recS... ebbD... ebbT 08:40:40:000 botT... actV... batL... irSE... chkS...\\ 
    \hline  
    RD 15 & recS... ebbD... ebbT 08:40:42:000 botT... actV... batL... irSE... chkS...\\ 
    \hline   
    RD 16 & recS... ebbD... ebbT 08:40:44:000 botT... actV... batL... irSE... chkS...\\ 
    \hline
    RD 17 & recS... ebbD... ebbT 08:40:45:000 botT... camF 01:00307200:... chkS...\\ 
    \hline
    \end{tabular}
    \caption{Example EBB illustrating variable frequency of RD sampling}
	\label{table9}
\end{table}





\end{document}